\documentclass[lettersize,journal]{IEEEtran}                                                          

\IEEEoverridecommandlockouts                                  

\usepackage{tabularray}
\usepackage{cite}
\usepackage{amsmath,amssymb,amsfonts}
\usepackage[]{footmisc}
\usepackage{xcolor}
\usepackage{graphicx}
\usepackage{textcomp}
\usepackage{psfrag}
\usepackage{epsfig}
\usepackage{subcaption}
\usepackage{float}
\usepackage{xcolor}
\usepackage{bm}
\usepackage{tabularx}
\usepackage{booktabs}
\usepackage{algorithm}
\usepackage{algpseudocode}
\usepackage[left=1.69cm,right=1.69cm,top=2.01cm,bottom=1.44cm]{geometry}

\newtheorem{mypro}{Proposition}

\newtheorem{myrem}{Remark}
\newtheorem{mylem}{Lemma}

\newcommand{\mR}{\mathcal{R}}
\newcommand{\ad}{\text{ad}}
\begin{document}

\title{\LARGE Lie-algebra Adaptive Tracking Control for Rigid Body Dynamics}

\author{Jiawei Tang, Shilei Li, and Ling Shi,~\IEEEmembership{Fellow,~IEEE}
\thanks{This work is supported by XXX.}
	\thanks{Jiawei Tang and Ling Shi are with the Department of Electronic and Computer Engineering, Hong Kong University of Science and Technology, Clear Water Bay, Hong Kong SAR (email: jtangas@connect.ust.hk; eesling@ust.hk).}
  \thanks{Shilei Li is with the School of Automation, Beijing Institute of Technology, China (email:  shileili@bit.edu.cn)}
	}

\maketitle

\begin{abstract}
Adaptive tracking control for rigid body dynamics is of critical importance in control and robotics, particularly for addressing uncertainties or variations in system model parameters. However, most existing adaptive control methods are designed for systems with states in vector spaces, often neglecting the manifold constraints inherent to robotic systems. In this work, we propose a novel Lie-algebra-based adaptive control method that leverages the intrinsic relationship between the special Euclidean group and its associated Lie algebra. By transforming the state space from the group manifold to a vector space, we derive a linear error dynamics model that decouples model parameters from the system state. This formulation enables the development of an adaptive optimal control method that is both geometrically consistent and computationally efficient. Extensive simulations demonstrate the effectiveness and efficiency of the proposed method. We have made our source code publicly available to the community to support further research and collaboration\footnote{https://github.com/Garyandtang/Lie-algebra-Adaptive-Control-Single-Rigid-Body-Dynamics}.
\end{abstract}
\IEEEpeerreviewmaketitle
\section{Introduction}
Tracking control of rigid body dynamics is a fundamental challenge in robotics, aerospace, and mechanical systems, with applications ranging from autonomous vehicles to robotic manipulators and spacecraft attitude control \cite{li2024prescribed, 10508075,10560008}. Precise trajectory tracking is essential for ensuring stability, safety, and performance in these systems. Rigid body dynamics, which describe the motion of object by external forces, are inherently nonlinear and complex, making control design a non-trivial task. Existing control methods often rely on local parameterizations, which can lead to singularities and performance degradation in globally defined tasks~\cite{pmlr-v155-song21a, ALTRO,9069467}. These challenges have motivated extensive research into advanced control techniques that leverage the geometric structure of rigid body dynamics to achieve robust and globally consistent performance. 

Rigid body motion is naturally described by the special Euclidean group $SE(n)$, which captures both translational and rotational dynamics in a unified framework \cite{murray2017mathematical}. The use of Lie group theory in control design provides a geometrically intuitive and globally consistent representation of rigid body dynamics, avoiding the singularities and ambiguities associated with local parameterizations such as Euler angles~\cite{micro-lie-theory}. Significant achievements have been reported in the literature~\cite{representation-free,RODRIGUEZCORTES2022105360}. Most existing methods addressing manifold constraint design the control objective on a Lie group and then derive the gradient for controller design. However, as shown by Teng et.al.~\cite{lie-algebra}, designing controllers in group space will face gradient vanishing issue, and slow down the convergence process. By leveraging the linear isomorphism between the Lie algebra and a vector space, they present a novel control framework on the general Lie group by designing the cost function in Lie algebra and show that setting the gradient of the cost function as the tracking error in the Lie algebra leads to a quadratic Lyapunov function that enables globally exponential convergence. Their findings benefit the development of the model predictive control (MPC)-based tracking control for different robotic systems, such as legged robot \cite{9981282}, marine vehicles \cite{teng-marien-vehicle}, and mobile robots \cite{GMPC}.

A critical challenge in rigid body control is the presence of model uncertainties, such as unknown or varying mass and inertia parameters. These uncertainties can significantly degrade the performance of model-based control methods, which often assume perfect knowledge of the system model. To handle model uncertainty, various adaptive control techniques have been proposed, including adaptive dynamics programming and adaptive optimal control. These methods rely on the persistence of excitation (PE) condition to ensure that the collected data is sufficiently rich for adaptive control. Theoretical works\cite{dean2018regret,frank-tnnls} demonstrate the convergence property of these online learning methods, facilitating their application to many different real-world platforms\cite{9497675,CHAKRABORTY2024103026,yang2022hierarchical}.

Even though adaptive control methods play important roles in many different applications, most existing adaptive control methods are designed for systems whose dynamics are naturally described in vector spaces, limiting their applicability to rigid body systems whose dynamics evolve on Lie groups. Their direct application to systems evolving on Lie groups is not straightforward. The nonlinear and non-Euclidean nature of Lie groups requires a fundamentally different approach to control design. For example, the state space of a rigid body is not a vector space but a manifold, and the dynamics are inherently coupled due to the group structure. This coupling complicates the design of adaptive control laws, as traditional methods for parameter estimation and control synthesis are not directly applicable. Moreover, the lack of a global coordinate system on Lie groups necessitates the use of geometric tools to formulate the control problem in a coordinate-invariant manner. To bridge this gap, we propose a novel Lie algebra adaptive tracking control framework that combines the geometric structure of Lie groups with adaptive control techniques. By formulating the error dynamics in the Lie algebra, we decouple the model parameters from the system state, enabling efficient reconstruction of uncertain parameters from collected data. The main contributions of this work are as follows:

\textcolor{black}{1) We derive a linear error dynamics model for rigid body tracking control by leveraging the relationship between the Lie group $SE(n)$ and its corresponding Lie algebra $\mathfrak{se}(n)$. This formulation allows us to transform the state space from group space into vector space and decouple the model parameters and the system state, facilitating the development of adaptive optimal control.}

\textcolor{black}{2) We develop a novel Lie-algebra adaptive control algorithm that reconstructs uncertain model parameters, such as mass and inertia, from collected system state and control input data. By decoupling the model parameters from the system state, the Lie algebra formulation provides a computationally efficient and geometrically intuitive framework for adaptive control of rigid body dynamics.}

\textcolor{black}{3) We demonstrate the effectiveness of the proposed method through numerical simulations, showing that it achieves precise trajectory tracking and robust performance in the presence of model uncertainties. Our code is available online to benefit the development of this research area.}

The remainder of this paper is organized as follows. In Section II, some preliminaries on the special Euclidean group and rigid body dynamics are given and the $SE(3)$ trajectory tracking control is introduced. In Section III,  the main results of our Lie-algebra adaptive control are presented. In Section IV, some simulations are provided to evaluate the effectiveness and efficiency of the proposed method. In Section V, the whole paper is concluded, and future work is discussed.

{\it Notations}:  The notation $\mathbb{R}^{n}$ denotes the set of real vectors with  $n$ elements. The notation $\mathbb{R}^{n\times m}$ denotes the set of the real matrix with  $n \times m$ elements.  The $i$-th element of a vector $ {x}\in \mathbb{R}^{n}$ is denoted by $x[i]$. The notations $I_{n}$ and $0_{n\times m}$ are denoted  the $n\times n$ identity matrix and $n\times m$ zero matrix, respectively. For notational clarity, the subscript will be dropped if the matrix dimension is clear. The norm $\|\cdot\|$in this paper is assumed to be Euclidean if not specified.

\section{Preliminaries}
In this section, some backgrounds on the special Euclidean group $SE(n)$ and the rigid body dynamics are provided. More details can be found in \cite{micro-lie-theory} for the Lie theory and \cite{murray2017mathematical} for the Lagrangian mechanism.

\subsection{Special Euclidean Group $SE(n)$}
Consider a rigid body located at $p \in \mathbb{R}^{n}$, and the orientation is described by a rotation matrix $\mathcal{R} \in SO(n)$, where the special orthogonal group $SO(n)$ is defined as
\begin{equation*}
    SO(n) = \{\mathcal{R} \in \mathbb{R}^{n \times n}~|~\mathcal{R}^{\top} \mathcal{R} = I, \det \mathcal{R} = 1\}.
\end{equation*}
The pose of the rigid body $X$ can be represented using the homogeneous representation, i.e., 
\begin{equation*}
    X = \begin{bmatrix}
        \mathcal{R} & p \\ 0_{1\times n} & 1
    \end{bmatrix}  \in \mathbb{R}^{(n+1)\times (n+1)}.
\end{equation*}
The combination of the set of all $X$ and the operation of matrix multiplication constitute a Lie group known as special Euclidean group $SE(n)$, i.e.,
\begin{equation*}
        SE(n)= \Big\{\begin{bmatrix}
        \mathcal{R} & p \\ 0_{1\times n} & 1
    \end{bmatrix} ~|~\mathcal{R}\in SO(n), ~p \in \mathbb{R}^n\Big\}. 
\end{equation*}
The tangent space at the identity of the special Euclidean group $SE(n)$ is defined as the Lie algebra $\mathfrak{se}(n)$. Moreover, there exists a linear isomorphism between the Lie algebra $\mathfrak{se}(n)$ and the vector space $\mathbb{R}^{p}$ (where $\dim \mathfrak{se}(n) = p$). For twist $\zeta := \begin{bmatrix}
     w \\ v
\end{bmatrix} \in \mathbb{R}^{p}$, where $w$ and $v$ are the angular and linear velocities in the body frame, the linear map to the Lie algebra $\mathfrak{se}(n)$ is denoted as $(\cdot)^{\wedge}$, i.e.,
\begin{align*}
    (\cdot)^{\wedge}&: \mathbb{R}^p \rightarrow \mathfrak{se}(n);~~~\zeta \rightarrow \zeta^{\wedge} = \begin{bmatrix}
        w^{\wedge} & v \\
         0 & 0
    \end{bmatrix}. \label{wedge}
\end{align*}
The inverse map of $(\cdot)^{\wedge}$ is denoted as $(\cdot)^{\vee}$. Besides, the matrix exponential map $\exp(\cdot)$ maps elements from Lie algebra to Lie group. For element $X \in SE(n)$, we have
\begin{align*}
    &\operatorname{exp}(\cdot): \mathfrak{se}(n) \rightarrow SE(n), ~\zeta^{\wedge}\rightarrow X = \exp(\zeta^{\wedge}),\\
     &\operatorname{log}(\cdot): SE(n) \rightarrow \mathfrak{se}(n), ~X\rightarrow \zeta^{\wedge} = \log(X) .
\end{align*}
where $\log(\cdot)$ is the inverse operation of $\exp(\cdot)$. Given a twist signal $\zeta(t)$, the motion of the rigid body can be expressed as follows
\begin{equation}\label{rigid-body-dyns}
    \dot{X}(t) = X(t) \zeta(t)^{\wedge},
\end{equation}
where $\dot{X}(t)$ is the first-order time-derivative of $X(t)$.

\subsection{Rigid Body Dynamics}
Consider a 3D rigid body whose configurations reside in the special Euclidean group $SE(3)$. The state is defined as $\{X, \zeta\} \in SE(3)\times \mathfrak{se}(3)$, where $X$ consists its center of mass (CoM) position $p \in \mathbb{R}^3$ and orientation $\mR \in SO(3)$ in the world frame and $\zeta$ consists its linear velocity $v \in \mathbb{R}^3$ and angular velocity $w \in \mathbb{R}^3$ in the body frame, i.e.,
\begin{align}
    X = \begin{bmatrix}
        R & p \\ 0 & 1
    \end{bmatrix} \in SE(3), ~~~\zeta =\begin{bmatrix}
        w \\ v
    \end{bmatrix} \in \mathbb{R}^6. \label{definition}
\end{align}
The control input $u \in \mathbb{R}^6$ on the rigid body is 
\begin{equation}
    u = \begin{bmatrix}
         \tau \\f
    \end{bmatrix} \in \mathbb{R}^6,
\end{equation}
where $f$ and $\tau$ are the force and torque applied at the CoM of the rigid body in the body frame. Following the Lie theory and Lagrangian mechanism, the continuous-time dynamics model of the rigid body is as follows:
\begin{equation} \label{rigid-body-dynamics}
    \begin{aligned}
    \dot{X}(t) &= X(t)\zeta(t)^{\wedge},\\
       \dot{\zeta}(t)&= J_b ^{-1}(\text{ad}_{\zeta}^{\top}J_b \zeta +u(t)) = f(\zeta(t), u(t)),
\end{aligned}
\end{equation}
where $\text{ad}_\zeta^{\top}$ is the coadjoint action map and $J_b$ is the generalized inertia matrix consisting the mass $m$ and the moment of inertia $I_b$ in the body frame, i.e.,
\begin{align*}
    \text{ad}^{\top}_{\zeta} = -\begin{bmatrix}
         w^{\wedge} & v^{\wedge} \\ 0 & w^{\wedge}
    \end{bmatrix}, ~~~ J_b = \begin{bmatrix}
        I_b & 0 \\ 0 & mI
    \end{bmatrix}.
\end{align*}
Ideally, these model parameters of the rigid body $m$, $I_b$ can be measured. However, in practical applications, measurement errors introduce model uncertainty, and only nominal parameter values are typically available, i.e.,
\begin{equation}\label{model uncertainty}
   \hat{m} = m  + \Delta_m, \; \hat{I_b} =I_b + \Delta_{I_b},
\end{equation}
where $\hat{(\cdot)}$ describes the nominal model parameter and $\Delta_{(\cdot)}$ describes the unknown additive model uncertainty.
\subsection{$SE(3)$ Trajectory Tracking}
Given a reference trajectory $\{X_d(t), \zeta_d(t), u_d(t)\}$ for tracking, we calculate the error state between the reference frame and the actual frame using group operations. the error state is defined as follows:
\begin{equation} \label{full-error-state}
    \begin{bmatrix}
        \Psi(t)\\ \delta\zeta(t)
    \end{bmatrix} = \begin{bmatrix}
        X_d(t)^{-1}X(t)\\ \zeta(t) - \zeta_d(t)
    \end{bmatrix},
\end{equation}
where $\Psi(t)$ is the pose error and $\delta \zeta(t)$ is the twist error. For $SE(3)$ trajectory tracking, we aim to design the control input $u$ to minimize the error state while subject to the rigid body dynamics \eqref{rigid-body-dynamics}. Besides, since the system is subject to model uncertainty as indicated in \eqref{model uncertainty}, we will present a Lie-algebra adaptive control method in the following section to address this uncertainty.
\section{Main Results}
In this section, we first derive error dynamics for the rigid body and linearize it using the relationship between the Lie group and Lie algebra. After that, a novel Lie-algebra adaptive optimal control is proposed to handle system dynamics' model uncertainty.
\subsection{Lie-algebra Error Dynamics}
\begin{mylem}
Given a feasible reference trajectory $\{X_d(t), \zeta_d(t), u_d(t)\}$ for rigid body \eqref{rigid-body-dynamics} to track, the error dynamics is as follows:
\begin{equation} \label{SE3-ERROR-Dynamics}
    \begin{aligned}
        \begin{bmatrix}
            \dot{\Psi}(t)\\ \dot{\delta \zeta}(t)
        \end{bmatrix} = \begin{bmatrix}
            \Psi(t) \zeta(t)^{\wedge}-\zeta_{d}(t)^{\wedge} \Psi(t)\\
            J_b^{-1} (\ad_{\zeta}^{\top}J_b\zeta(t) + u(t) -\ad_{\zeta_d}^{\top}J_b\zeta_d(t) - u_d(t))
        \end{bmatrix}
    \end{aligned}.
\end{equation}
\noindent \textit{Proof:} Taking derivative of the pose error in \eqref{full-error-state}, we have
\begin{equation}\label{pose-error-dynamics}
    \begin{aligned}
\dot{\Psi}(t)&=\frac{d}{d t}\left(X_{d}(t)^{-1}\right) X(t)+X_{d}(t)^{-1} \frac{d}{d t} X(t) \\
& =X_{d}(t)^{-1} \frac{d}{d t} X(t)-X_{d}(t)^{-1} \frac{d}{d t}\left(X_{d}(t)\right) X_{d}(t)^{-1} X(t) \\
& =X_{d}(t)^{-1} X(t) \zeta(t)^{\wedge}-X_{d}(t)^{-1} X_{d}(t) \zeta_{d}(t)^{\wedge} X_{d}(t)^{-1} X(t) \\
& =\Psi(t) \zeta(t)^{\wedge}-\zeta_{d}(t)^{\wedge} \Psi(t),
\end{aligned}
\end{equation}
Similarly, the twist error dynamics is as follows:
\begin{equation*}
\begin{aligned}
    \dot{\delta \zeta}(t) &= \dot{\zeta}(t) - \dot{\zeta}_d(t)\\
    &= J_b^{-1} (\ad_{\zeta}^{\top}J_b\zeta(t) + u(t) -\ad_{\zeta_d}^{\top}J_b\zeta_d(t) - u_d(t)).\\
\end{aligned}
\end{equation*}
Combining the pose error and twist error together, the overall error dynamics model is as follows:
\begin{equation*}
    \begin{aligned}
        \begin{bmatrix}
            \dot{\Psi}(t)\\ \dot{\delta \zeta}(t)
        \end{bmatrix} = \begin{bmatrix}
            \Psi(t) \zeta(t)^{\wedge}-\zeta_{d}(t)^{\wedge} \Psi(t)\\
            J_b^{-1} (\ad_{\zeta}^{\top}J_b\zeta(t) + u(t) -\ad_{\zeta_d}^{\top}J_b\zeta_d(t) - u_d(t))
        \end{bmatrix}.
    \end{aligned}
\end{equation*}
The proof is completed. \hfill $\square$ 
\end{mylem}

The aforementioned results characterize the tracking dynamics within the special Euclidean group $SE(3)$. Noticing that the pose error $\Psi(t)$ remains in $SE(3)$. In the subsequent discussion, we will leverage the relationship between $SE(3)$ and its corresponding Lie algebra $\mathfrak{se}(3)$ to transform the state space from the group space to a vector space. This transformation will facilitate the linearization of the error dynamics.

\begin{mypro}
Given the operation point $\{\zeta_d, {u}_d\}$,  define $\psi(t)^{\wedge} := \log(\Psi(t))$ as the pose error in Lie algebra $\mathfrak{se}(3)$ and $\delta u(t) := u(t) - u_d(t)$ as the perturbed control input, the linear rigid body error dynamics in vector space is as follows:
\begin{equation}
    \begin{bmatrix}
    \dot{\psi}(t)\\ \dot{\delta \zeta}(t)
\end{bmatrix} = \begin{bmatrix}
        -\ad_{\zeta_d} & I \\ 0 & \Gamma_{\zeta_d}
    \end{bmatrix}\begin{bmatrix}
    \psi(t)\\ \delta \zeta(t)
\end{bmatrix} + \begin{bmatrix}
    0 \\ J_b^{-1}
\end{bmatrix}\delta u(t),
 \end{equation}
 where
 \begin{equation*}
    \begin{aligned}
        \Gamma_{\zeta_d}= J_b^{-1}\ad_{\zeta_d}^{\top}J_b + J_b^{-1}\begin{bmatrix}
            (I_b w_d)^{\wedge} & m v_d^{\wedge}\\
            m v_d^{\wedge} & 0
        \end{bmatrix}.
    \end{aligned}
\end{equation*}
\textit{Proof:} Since $\Psi(t) = \exp(\psi(t)^{\wedge})$, taking first-order Taylor approximation of the matrix exponential map $\exp(\cdot)$, we have
\begin{equation}\label{psi_exp1}
    \Psi(t)  =\exp(\psi(t)^{\wedge})= I + \sum_{i=1}^{\infty}\frac{1}{i!}(\psi(t)^{\wedge})^i \approx I + \psi(t)^{\wedge}.
\end{equation}
Substituting \eqref{psi_exp1} to \eqref{pose-error-dynamics}, we have
\begin{equation}\label{linear error se3 dynamics}
    \dot{\psi}(t)^{\wedge}  \approx \zeta(t)^{\wedge} - \zeta_d(t)^{\wedge}+\psi(t)^{\wedge}\zeta_d(t)^{\wedge}-\zeta_{d}(t)^{\wedge}\psi(t)^{\wedge}.
\end{equation}
Following the adjoint map in Lie Algebra, i.e.,
\begin{equation*}
    \ad_{\phi}(\eta) = [\phi^{\wedge}, \eta^{\wedge}],
\end{equation*}
and taking $(\cdot)^{\vee}$ operation on both side of \eqref{linear error se3 dynamics}, we have the pose error dynamics in vector space as follows:
\begin{equation} \label{linear perturbed configuration dyna}
        \dot{\psi}(t) =-\ad_{\zeta_d}\psi(t) + \zeta(t) - \zeta_d(t).
\end{equation}
Similarly, taking a first-order Taylor approximation of the twist error dynamics with respect to the operating point $\{\zeta_d, u_d\}$ and dropping the time notation for clarity, we have:
\begin{equation*}
    \dot{\zeta} \approx f(\zeta_d, u_d)+ \frac{\partial f(\zeta, u)}{\partial \zeta}|_{\zeta_d, u_d} (\zeta - \zeta_d)+ \frac{\partial f(\zeta, u)}{\partial u}|_{\zeta_d, u_d}(u - u_d),
\end{equation*}
where
\begin{equation*}
    \begin{aligned}
        f(\zeta_d, u_d) &= \dot{\zeta}_d,\\
        \frac{\partial f(\zeta, u)}{\partial \zeta}|_{\zeta_d, u_d} &= J_b^{-1}\ad_{\zeta_d}^{\top}J_b + J_b^{-1}\begin{bmatrix}
            (I_b w_d)^{\wedge} & m v_d^{\wedge}\\
            m v_d^{\wedge} & 0
        \end{bmatrix}  = \Gamma_{\zeta_d},\\
        \frac{\partial f(\zeta, u)}{\partial u}|_{\zeta_d, u_d} &= J_b^{-1}.
    \end{aligned}
\end{equation*}
 Define the perturbed control input $\delta u = u - u_d$. The linear twist error dynamics is as follows:
\begin{equation} \label{linearized perturbed velocity dynamics}
    \dot{\delta \zeta} = \Gamma_{\zeta_d} \delta\zeta + J_b^{-1} \delta u.
\end{equation}
Combining the linear pose error dynamics and twist error dynamics together, the overall linear error dynamics is
\begin{equation}
    \begin{bmatrix}
    \dot{\psi}(t)\\ \dot{\delta \zeta}(t)
\end{bmatrix} = \begin{bmatrix}
        -\ad_{\zeta_d} & I \\ 0 & \Gamma_{\zeta_d}
    \end{bmatrix}\begin{bmatrix}
    \psi(t)\\ \delta \zeta(t)
\end{bmatrix} + \begin{bmatrix}
    0 \\ J_b^{-1}
\end{bmatrix}\delta u(t).
 \end{equation}
 The proof is completed. \hfill $\square$
\end{mypro}
\begin{figure*}[t]
    \centering
    \begin{subfigure}{0.5\textwidth}
        \includegraphics[width=\textwidth]{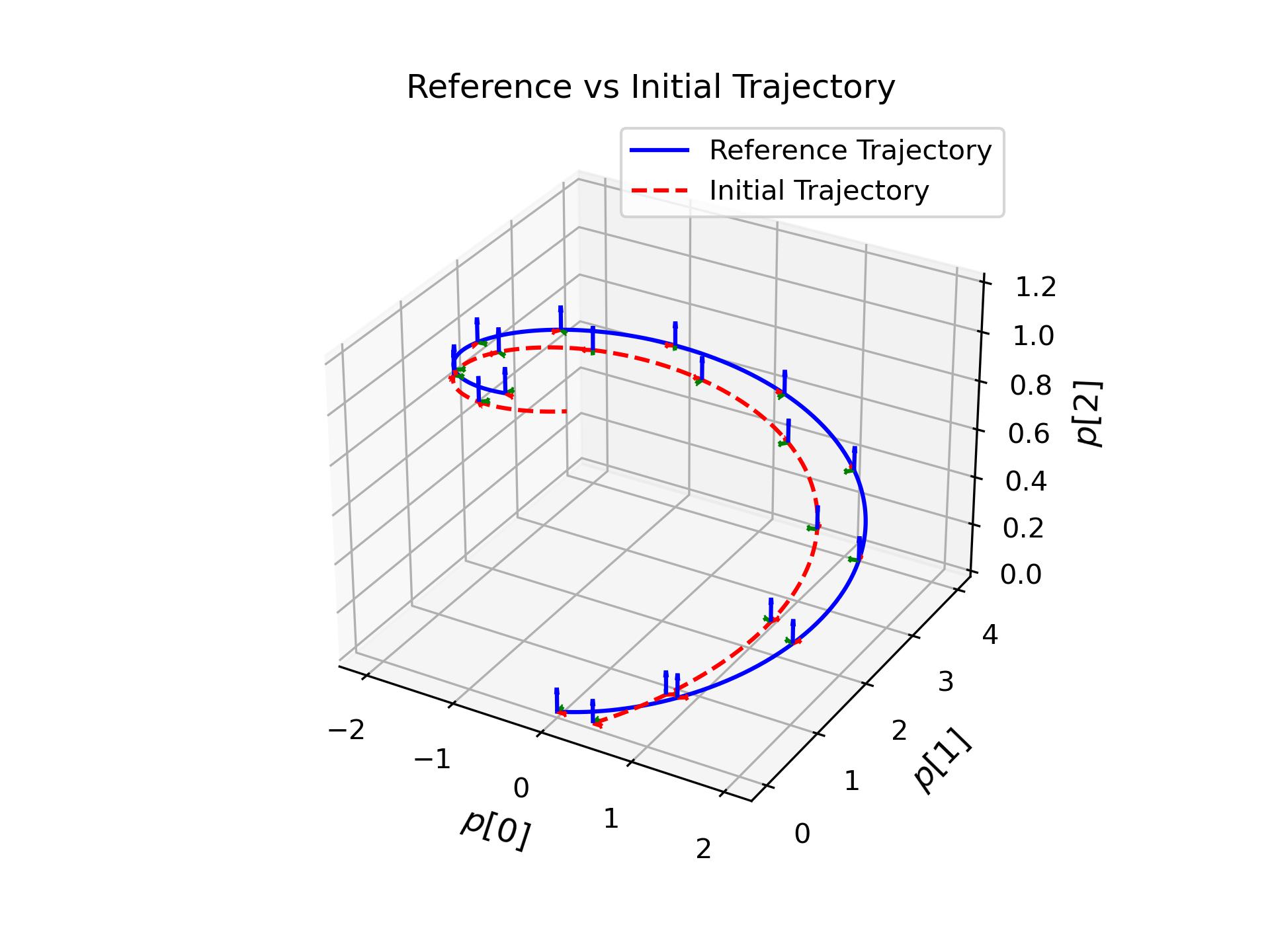}
        
        \caption{Trajectories with initial parameters.}
       
    \end{subfigure}%
    \begin{subfigure}{0.5\textwidth}
        \includegraphics[width=\textwidth]{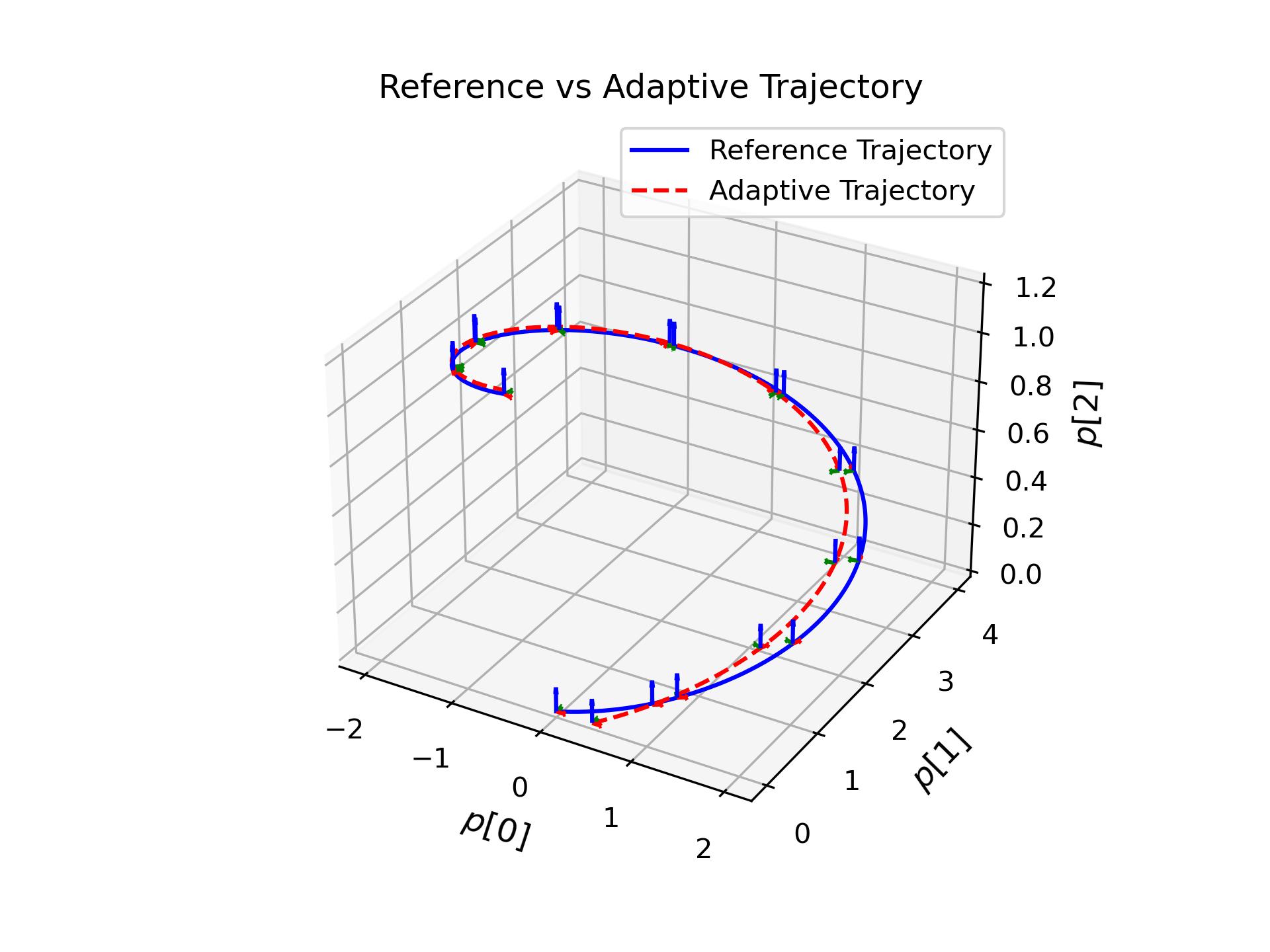}
        \caption{Trajectories with adaptive parameters.}
      
    \end{subfigure}%

    \caption{Trajectory tracking comparison between initial model parameters and adaptive model parameters. The initial position is $\begin{bmatrix}
        0.4 & 0 & 0
    \end{bmatrix}^{\top}$, the initial orientation is $I$, the initial angular velocity and linear velocity are $\begin{bmatrix}
        0 & 0 & 0
    \end{bmatrix}^{\top}$. The dataset size for adaptive control is 1500 in this example.}
    \label{fig:initial-adaptive-comparison}
\end{figure*}

Different from \eqref{SE3-ERROR-Dynamics}, \textit{Proposition 1} characterizes the tracking dynamics in the vector space. Since both the state and control input are in vector space, it allows us to leverage algebraic methods to design an objective function for tracking control. Moreover, it is noteworthy that due to the symmetric property of the Lie group, the state transition matrix and control matrix are error state independent (depending only on the reference trajectory $\zeta_d$, $u_d$ and model parameters $I_b$, $m$). This property motivates the design of a Lie-algebra adaptive optimal control algorithm to address model uncertainty.
\subsection{Lie-algebra Adaptive Control}
Define $x := \begin{bmatrix}
    \psi \\ \delta \zeta
\end{bmatrix}$ and follow the standard method for discretizing continuous-time ordinary differential equation\cite{lewis2012optimal}, the corresponding discrete-time linear system is 
\begin{equation}\label{dt-linear}
        x_{k+1} =A_{\zeta_d}x_k + B\delta u, 
\end{equation}
where 
\begin{equation*}
    A_{\zeta_d} = I +  \begin{bmatrix}
        -\ad_{\zeta_d} & I \\ 0 & \Gamma_{\zeta_d}
    \end{bmatrix}\Delta t,~ B = \begin{bmatrix}
    0 \\ J_b^{-1}
\end{bmatrix}\Delta t, ~ \Delta t = t_k - t_{k-1}.
\end{equation*}

\textcolor{black}{It is easy to verify that the controllability Grammian of \eqref{dt-linear} is full rank. Hence, system \eqref{dt-linear} is controllable.} Since the state and control input space are in vector space, given semi-positive definite state cost matrix $Q \in \mathbb{R}^{12 \times 12}$ and the positive definite control cost matrix $R \in \mathbb{R}^{6 \times 6}$, the optimal tracking control problem is defined as follows:

\noindent\textit{{Problem 1}: (Lie-algebra Optimal Tracking Control)}
\begin{subequations}
\begin{align}
    \min_{u} &~\sum_{k=0}^{T}x_k^{\top}Q x_k+ \delta{u}_k^{\top}R\delta{u}_k\\
    \text{s.t.} &~x_{k+1} =A_{\zeta_d}x_k + B\delta{u}_k.\label{p1-constraint}
\end{align}
\end{subequations}

Given complete knowledge of the system model $(A_{\zeta_d}, B)$, \textit{Problem 1} can be effectively addressed by solving the corresponding discrete algebraic Riccati equation (DARE), 
\begin{equation*}
P^{*}={A_{\zeta_d}}^{\top} P^{*} {A_{\zeta_d}}+Q-{A_{\zeta_d}}^{\top} P^{*} {{B}}\left({{B}}^{\top} P^{*} {{B}}+R\right)^{-1} {{B}}^{\top} P^{*} {A_{\zeta_d}},
\end{equation*}
and yields an optimal state feedback control policy $\delta u_k = K x_k$. In what follows, we will utilize the collected data on system states and control inputs to adapt to the model uncertainty present in $(A_{\zeta_d}, B)$. 

Given a set of collected data of system state and control input $\{x_0, \delta u_0, x_1, \delta u_1, \ldots, x_{N}\}$, we define the dataset $(X_{-}, U, X_{+})$ as follows
\begin{equation}\label{dataset}
    X_{-} = \begin{bmatrix}
        x_{0}^{\top} \\ \vdots \\ x_{N-1}^{\top}
    \end{bmatrix}, 
    U = \begin{bmatrix}
        \delta u_{0}^{\top} \\ \vdots \\ \delta u_{N-1}^{\top}
    \end{bmatrix},
    X_{+} = \begin{bmatrix}
        x_{1}^{\top} \\ \vdots \\ x_{N}^{\top}
    \end{bmatrix}.
\end{equation}
Based on \eqref{p1-constraint}, we have 
\begin{equation}\label{eq1}
     \begin{bmatrix}
        X & U
    \end{bmatrix}
    \begin{bmatrix}
        A^{\top} \\ B^{\top}
    \end{bmatrix} =  X_{+}.
\end{equation}
Equation \eqref{eq1} formulates a least squares problem for $\begin{bmatrix}
        A^{\top} \\ B^{\top}
    \end{bmatrix}$. To mitigate the risk of overfitting and control model complexity~\cite{ljung2010perspectives}, we introduce a regularization term similar to those employed in~\cite{dean2018regret,abeille2017thompson}:

\begin{equation} \label{final-least-square}
    (\begin{bmatrix}
        X^{\top} \\ U^{\top}
    \end{bmatrix}
    \begin{bmatrix}
        X & U
    \end{bmatrix} + \lambda I)
    \begin{bmatrix}
        A^{\top} \\ B^{\top}
    \end{bmatrix} = 
    \begin{bmatrix}
        X^{\top} \\ U^{\top}
    \end{bmatrix} X_{+},
\end{equation}
where $\lambda > 0$ is the regulation term. 
By solving the least squares problem for $(A, B)$, we can reconstruct the model parameters $(I_b, m)$ from dataset $(X_{-}, U, X_{+})$. The final algorithm is summarized as follows.
\begin{algorithm}[h]
\caption{Lie-algebra Adaptive Control Algorithm}\label{alg: lie-algebra-adaptive}
\begin{algorithmic}
\Require $Q \geq 0$, $R > 0$, $\lambda > 0$, $\zeta_d$, $u_d$
\Require Initialize $K$
\State Let ${u}_k = K x_k +u_d+ \gamma_k$ // \textit{remark 1}  
\State Execute $u$ for $N$ steps and collect $x_0,u_0,x_1,u_1,\ldots,x_{N}$
\State Obtain dataset by \eqref{dataset}
\State Formulate a least-square problem for (A, B) by \eqref{final-least-square}
\State Obtain (A, B) by solving the corresponding problem.
\State Reconstruct ($I_b$, $m$) by \eqref{dt-linear}

\State \Return $I_b$, $m$
\end{algorithmic}
\end{algorithm}
\begin{myrem}
In adaptive control, ensuring the persistence of excitation (PE) condition \cite{rl-fbc} is essential to guarantee that the collected data is sufficiently rich for controller design. To satisfy the PE condition, an exploration noise term, denoted by $\gamma$ is incorporated into the control input. Similarly to \cite{dean2018regret,abeille2017thompson}, this noise signal is randomly sampled from a zero-mean normal distribution in our implementation.
\end{myrem}
\begin{myrem}
    Unlike formulating the optimal control problem using the original rigid body dynamics \eqref{rigid-body-dynamics}, the Lie-algebra formulation enables the decoupling of model parameters and system state. This decoupling facilitates the efficient reconstruction of model parameters from the collected data.  
\end{myrem}

\textbf{Algorithm \ref{alg: lie-algebra-adaptive}} presents an adaptive approach to address unknown model uncertainties by leveraging collected system state and control input data, along with a predefined reference trajectory. The optimal control policy can be obtained by solving the DARE using the reconstructed model parameters. For time-varying tracking control problems, the reconstructed model parameters can be utilized in a time-varying linear quadratic regulator (TVLQR) or model predictive control (MPC) framework. Given the quadratic objective function and linear constraints, the MPC controller can be efficiently implemented using standard quadratic programming (QP) solvers.

\section{Simulations}

In this section, we apply our Lie-algebra adaptive control method to a fully actuated three-dimensional rigid body system, as described by the dynamics in \eqref{rigid-body-dynamics}, for trajectory tracking. The reference angular velocity $w_d$
  and linear velocity $v_d$ are set to
  \begin{equation*}
      w_d = \begin{bmatrix}
    0 & 0 & 1
\end{bmatrix}^{\top}, \;
v_d = \begin{bmatrix}
    2 & 0 & 0.2
\end{bmatrix}^{\top}.
  \end{equation*}
The reference control input $u_d$ is set to
\begin{equation*}
    u_d = \begin{bmatrix}
    0_{1 \times 3} & m (w_d^{\wedge} v_d)^{\top} 
\end{bmatrix}^{\top},
\end{equation*}
which is designed to counteract the Coriolis force. 
The reference pose $X_d$ is obtained by integrating the reference twist and control from the initial pose at the origin, i.e., $X_0 = I$.

The inertia matrix $I_b$ and mass $m$ are
\begin{equation*}
    I_b = \begin{bmatrix}
        1 & 0.2 & 0.1\\
        0.2 & 1 & 0.2 \\
        0.1 & 0.2 & 1
    \end{bmatrix}, \; m = 2,
\end{equation*}
which are unknown to the controller and only the measured model parameters $\hat{I_b}$ and $\hat{m}$ are available. The additive uncertainty in the inertia matrix is randomly selected from a set of positive definite matrices, while the additive mass uncertainty is randomly sampled from a uniform distribution.

\begin{figure}[t]
      \centering
	   \begin{subfigure}{0.9\linewidth} \label{evolution I}
		\includegraphics[width=\linewidth]{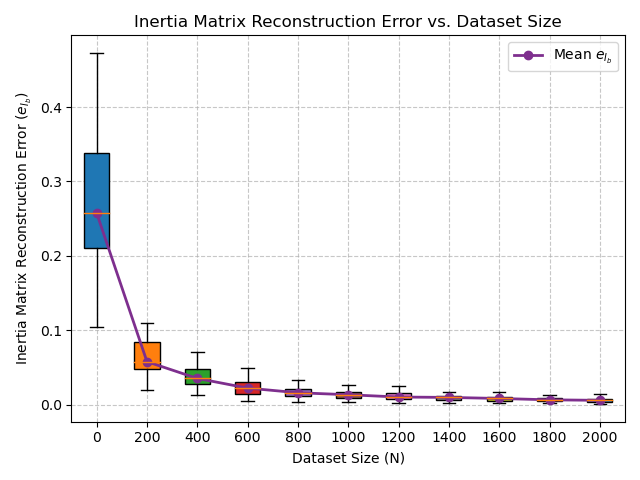}
		\caption{Evolution of inertia matrix reconstruction error $e_{I_b}$.}

	   \end{subfigure}
    \quad
	   \begin{subfigure}{0.9\linewidth} \label{evolution m}
		\includegraphics[width=\linewidth]{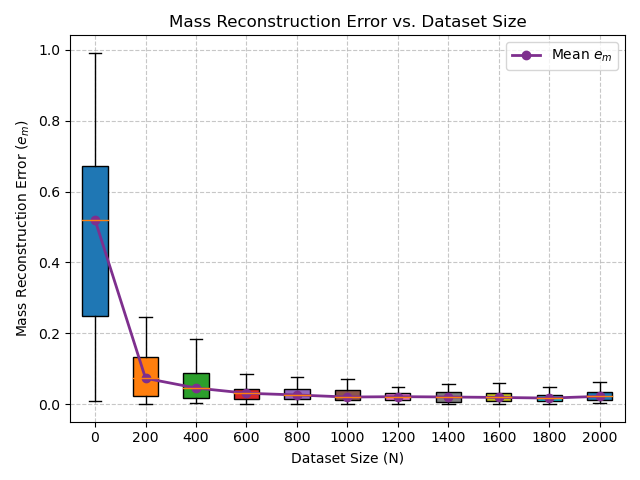}
		\caption{Evolution of mass reconstruction error $e_{m}$.}

	    \end{subfigure}
     \caption{Evolution of model parameters in terms of dataset size $N$.}
     \label{fig:evolution of I m}
    \end{figure}

\begin{figure}[t]
	\centering
	\includegraphics[width=8cm]{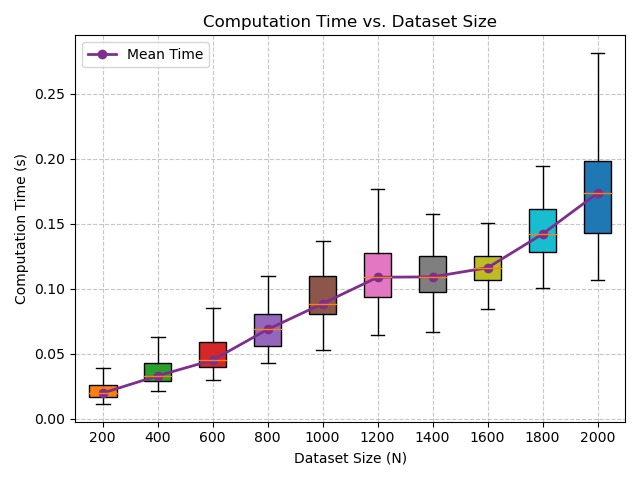}
	\caption{Evolution of computation time in terms of dataset size.}
	\label{fig:evolution compilation time}
\end{figure}

The efficacy of \textbf{Algorithm 1} is evaluated through Monte Carlo experiments. The reconstruction error for the matrix is assessed using the Frobenius norm, while the vector is quantified using the \( \ell_2 \)-norm. In the experiments, 50 initial measured parameters are randomly selected, and the dataset size is varied as \( N = \{200, 400, 600, 800, 1000, 1200, 1400, 1600, 1800, 2000\} \). Figure~\ref{fig:initial-adaptive-comparison} illustrates the difference between the trajectory generated using the initial model parameters and the one generated using the reconstructed parameters. Figure~\ref{fig:evolution of I m}  depicts the evolution of \( e_{I_b} \) and \( e_{m} \), respectively, as a function of dataset size. The results demonstrate that both \( e_{I_b} \) and \( e_{m} \) decrease as the dataset size increases, enabling the system to track the reference trajectory with greater precision. We also evaluate the run time performance of our method. \textcolor{black}{Figure~\ref{fig:evolution compilation time} shows the computational efficiency of \textbf{Algorithm 1} across varying dataset sizes. We observe that the run time increases linearly with $N$, indicating a consistent performance that scales with the amount of data processed. Moreover, when the dataset size $N=2000$, the average run time is around $0.17s$, which is very fast ensuring that the algorithm is practical for real-time applications. One thing that should be mentioned is that all the experiments are conducted on a laptop computer equipped with an Apple M3 chip.}

We also evaluate the tracking performance of the proposed method. In the experiments, we compare our method with two online learning methods, named \textit{Optimization in the Face of Uncertainty-based LQR} (OFU) \cite{dean2018regret} and \textit{Thompson Sampling LQR} (TS)\cite{abeille2017thompson}. To make the comparison fair, the optimal control formulation for these two baseline methods is also based on \textit{Problem 1}. Table I summarizes the tracking performance of our method and two baseline methods. The symbols $e_p$, $e_\mathcal{R}$, $e_w$, and $e_v$ denote the tracking error of position, rotation, angular velocity and linear velocity, respectively. It is easy to observe that the proposed Lie-algebra adaptive control method achieves significantly smaller tracking errors compared to the OFU and TS-based controllers. Specifically, the controllers created by the OFU and TS methods exhibit large tracking errors. This is likely due to their reliance on vector space formulations, which fail to account for the geometric structure and manifold constraints of rigid body dynamics. In contrast, our method, which leverages the Lie algebra framework, ensures a more accurate and consistent representation of the system, leading to superior tracking performance.
\section{Conclusion}
This paper addressed the challenge of trajectory tracking control for single rigid body dynamics in the presence of model uncertainties. Building on the geometric properties of Lie algebras and recent advances in adaptive control, we developed a novel Lie-algebra-based tracking control method. This method leverages system state and control input data to adaptively improve tracking performance, offering a computationally efficient and robust solution to handle uncertainties in the dynamics model. Experimental validation demonstrated the effectiveness of the proposed method, showcasing its ability to achieve high tracking accuracy while maintaining time and data efficiency. By unifying principles from geometric control and adaptive techniques, our work provides a foundation for applying Lie-algebraic methods to a broader class of robotic systems. Future work will explore the integration of the proposed Lie-algebra tracking control framework with advanced machine learning techniques, such as deep reinforcement learning, to further enhance adaptability in highly uncertain and dynamic environments. 

\begin{table}[t]
\begin{center}\label{table:tracking error comparison}
\caption{Tracking Error Comparison}
\resizebox{\linewidth}{!}{
\begin{tabular}{|l||l|l|l|l|}
\hline
{Method $\backslash$ Metric} & {$e_p$} & {$e_\mathcal{R}$} & {$e_w$}& {$e_v$} \\ \hline
\textit{OFU} \cite{dean2018regret}      &     0.759        & 0.063 & 0.042 & 0.637 \\ \hline
\textit{TS  } \cite{abeille2017thompson}     &   0.505       & 0.64 & 0.372 & 0.516 \\ \hline
\textit{Our}       &     \textbf{0.011}        & \textbf{0.013} & \textbf{0.016 }& \textbf{0.029}  \\ \hline
\end{tabular}
}
\end{center}
\end{table}

\bibliographystyle{IEEEtran}
\bibliography{reference}

\begin{thebibliography}{10}
\providecommand{\url}[1]{#1}
\csname url@samestyle\endcsname
\providecommand{\newblock}{\relax}
\providecommand{\bibinfo}[2]{#2}
\providecommand{\BIBentrySTDinterwordspacing}{\spaceskip=0pt\relax}
\providecommand{\BIBentryALTinterwordstretchfactor}{4}
\providecommand{\BIBentryALTinterwordspacing}{\spaceskip=\fontdimen2\font plus
\BIBentryALTinterwordstretchfactor\fontdimen3\font minus \fontdimen4\font\relax}
\providecommand{\BIBforeignlanguage}[2]{{%
\expandafter\ifx\csname l@#1\endcsname\relax
\typeout{** WARNING: IEEEtran.bst: No hyphenation pattern has been}%
\typeout{** loaded for the language `#1'. Using the pattern for}%
\typeout{** the default language instead.}%
\else
\language=\csname l@#1\endcsname
\fi
#2}}
\providecommand{\BIBdecl}{\relax}
\BIBdecl

\bibitem{li2024prescribed}
J.~Li and S.~Song, ``Prescribed-time collision-free trajectory tracking control for spacecraft with position-only measurements,'' \emph{IEEE Transactions on Aerospace and Electronic Systems}, 2024.

\bibitem{10508075}
M.~Di~Ferdinando, S.~Di~Gennaro, D.~Bianchi, and P.~Pepe, ``On robust quantized sampled-data tracking control of nonlinear systems,'' \emph{IEEE Transactions on Automatic Control}, vol.~69, no.~10, pp. 7120--7127, 2024.

\bibitem{10560008}
S.~Zhang, S.~Ifqir, and V.~Puig, ``Linear quadratic zonotopic control of switched systems: Application to autonomous vehicle path-tracking,'' \emph{IEEE Control Systems Letters}, vol.~8, pp. 1895--1900, 2024.

\bibitem{pmlr-v155-song21a}
Y.~Song, S.~Naji, E.~Kaufmann, A.~Loquercio, and D.~Scaramuzza, ``Flightmare: A flexible quadrotor simulator,'' in \emph{Proceedings of the 2020 Conference on Robot Learning}, ser. Proceedings of Machine Learning Research, J.~Kober, F.~Ramos, and C.~Tomlin, Eds., vol. 155.\hskip 1em plus 0.5em minus 0.4em\relax PMLR, 16--18 Nov 2021, pp. 1147--1157.

\bibitem{ALTRO}
T.~A. Howell, B.~E. Jackson, and Z.~Manchester, ``{ALTRO}: A fast solver for constrained trajectory optimization,'' in \emph{IEEE/RSJ International Conference on Intelligent Robots and Systems (IROS)}, 2019, pp. 7674--7679.

\bibitem{9069467}
H.~Liang, H.~Li, and D.~Xu, ``Nonlinear model predictive trajectory tracking control of underactuated marine vehicles: Theory and experiment,'' \emph{IEEE Transactions on Industrial Electronics}, vol.~68, no.~5, pp. 4238--4248, 2021.

\bibitem{murray2017mathematical}
R.~M. Murray, Z.~Li, and S.~S. Sastry, \emph{A mathematical introduction to robotic manipulation}.\hskip 1em plus 0.5em minus 0.4em\relax CRC press, 2017.

\bibitem{micro-lie-theory}
J.~Sol{\`a}, J.~Deray, and D.~Atchuthan, ``A micro lie theory for state estimation in robotics,'' \emph{ArXiv}, vol. abs/1812.01537, 2018.

\bibitem{representation-free}
Y.~Ding, A.~Pandala, C.~Li, Y.-H. Shin, and H.-W. Park, ``Representation-free model predictive control for dynamic motions in quadrupeds,'' \emph{IEEE Transactions on Robotics}, vol.~37, no.~4, pp. 1154--1171, 2021.

\bibitem{RODRIGUEZCORTES2022105360}
H.~Rodríguez-Cortés and M.~Velasco-Villa, ``A new geometric trajectory tracking controller for the unicycle mobile robot,'' \emph{Systems \& Control Letters}, vol. 168, p. 105360, 2022.

\bibitem{lie-algebra}
S.~Teng, W.~Clark, A.~Bloch, R.~Vasudevan, and M.~Ghaffari, ``Lie algebraic cost function design for control on {Lie} groups,'' in \emph{IEEE 61st Conference on Decision and Control (CDC)}, 2022, pp. 1867--1874.

\bibitem{9981282}
S.~Teng, D.~Chen, W.~Clark, and M.~Ghaffari, ``An error-state model predictive control on connected matrix lie groups for legged robot control,'' in \emph{IEEE/RSJ International Conference on Intelligent Robots and Systems (IROS)}, 2022, pp. 8850--8857.

\bibitem{teng-marien-vehicle}
J.~Jang, S.~Teng, and M.~Ghaffari, ``Convex geometric trajectory tracking using lie algebraic mpc for autonomous marine vehicles,'' \emph{IEEE Robotics and Automation Letters}, vol.~8, no.~12, pp. 8374--8381, 2023.

\bibitem{GMPC}
J.~Tang, S.~Wu, B.~Lan, Y.~Dong, Y.~Jin, G.~Tian, W.-A. Zhang, and L.~Shi, ``{GMPC}: Geometric model predictive control for wheeled mobile robot trajectory tracking,'' \emph{IEEE Robotics and Automation Letters}, pp. 1--8, 2024.

\bibitem{dean2018regret}
S.~Dean, H.~Mania, N.~Matni, B.~Recht, and S.~Tu, ``Regret bounds for robust adaptive control of the linear quadratic regulator,'' \emph{Advances in Neural Information Processing Systems}, vol.~31, 2018.

\bibitem{frank-tnnls}
B.~Kiumarsi, K.~G. Vamvoudakis, H.~Modares, and F.~L. Lewis, ``Optimal and autonomous control using reinforcement learning: A survey,'' \emph{IEEE Transactions on Neural Networks and Learning Systems}, vol.~29, no.~6, pp. 2042--2062, 2018.

\bibitem{9497675}
L.~Cui, S.~Wang, J.~Zhang, D.~Zhang, J.~Lai, Y.~Zheng, Z.~Zhang, and Z.-P. Jiang, ``Learning-based balance control of wheel-legged robots,'' \emph{IEEE Robotics and Automation Letters}, vol.~6, no.~4, pp. 7667--7674, 2021.

\bibitem{CHAKRABORTY2024103026}
S.~Chakraborty, L.~Cui, K.~Ozbay, and Z.-P. Jiang, ``Automated lane changing control in mixed traffic: An adaptive dynamic programming approach,'' \emph{Transportation Research Part B: Methodological}, vol. 187, p. 103026, 2024.

\bibitem{yang2022hierarchical}
B.~Yang, Y.~Lu, X.~Yang, and Y.~Mo, ``A hierarchical control framework for drift maneuvering of autonomous vehicles,'' in \emph{IEEE International Conference on Robotics and Automation (ICRA)}, 2022, pp. 1387--1393.

\bibitem{lewis2012optimal}
F.~L. Lewis, D.~Vrabie, and V.~L. Syrmos, \emph{Optimal Control}.\hskip 1em plus 0.5em minus 0.4em\relax John Wiley \& Sons, 2012.

\bibitem{ljung2010perspectives}
L.~Ljung, ``Perspectives on system identification,'' \emph{Annual Reviews in Control}, vol.~34, no.~1, pp. 1--12, 2010.

\bibitem{abeille2017thompson}
M.~Abeille and A.~Lazaric, ``Thompson sampling for linear-quadratic control problems,'' in \emph{Artificial intelligence and statistics}.\hskip 1em plus 0.5em minus 0.4em\relax PMLR, 2017, pp. 1246--1254.

\bibitem{rl-fbc}
F.~L. Lewis and D.~Vrabie, ``Reinforcement learning and adaptive dynamic programming for feedback control,'' \emph{IEEE Circuits and Systems Magazine}, vol.~9, no.~3, pp. 32--50, 2009.

\end{thebibliography}
\end{document}